\documentclass[11pt]{article}

\usepackage[margin=1in]{geometry}
\usepackage{amsmath,amssymb}
\usepackage{booktabs}
\usepackage{graphicx}
\usepackage{caption}
\usepackage{subcaption}
\usepackage{hyperref}
\usepackage{xcolor}

\hypersetup{
  colorlinks=true,
  linkcolor=blue!60!black,
  citecolor=blue!60!black,
  urlcolor=blue!60!black
}
\newcommand{\best}[1]{{\boldmath\textbf{#1}}}

\title{Gauging, Measuring, and Controlling Critic Complexity in Actor-Critic Reinforcement Learning}
\author{Konstantin Garbers\\
Peking University\\
\texttt{konstantin.garbers25@stu.pku.edu.cn}}
\date{\today}

\begin{document}
\maketitle

\begin{abstract}
Actor-critic methods depend on learned critics, but critic quality is often
evaluated only indirectly through return, temporal-difference error, or value
loss. Critic complexity is introduced as an additional diagnostic and
intervention dimension for actor-critic reinforcement learning. The analysis
uses spectral effective-rank entropy, a rank-like summary of the
singular-value distributions of critic weight matrices, to assess critic
model complexity. Across TD3 and PPO experiments, critic complexity is
tracked together with return and Monte Carlo value-estimation bias. The
results show that critic complexity is measurable throughout training and is
systematically associated with training behavior, while also making clear
that the relationship is heterogeneous across algorithms, tasks, and
hyperparameters. A direct complexity-control intervention is then evaluated
by adding a spectral-entropy penalty to the critic loss. This intervention
reliably changes the targeted spectral quantity, demonstrating that critic
complexity can be controlled rather than only observed. Return effects are
treated as task-dependent evidence rather than as a general performance
claim, because overall complexity-control results vary.
\end{abstract}

\section{Introduction}

Actor-critic reinforcement learning depends on learned critics to estimate values and guide policy improvement. This makes the critic a central source of both progress and failure. If the critic overestimates certain states or actions, propagates bootstrapping errors, or learns an unnecessarily irregular value function, the actor may optimize against a distorted objective.

Much of actor-critic research is therefore concerned with improving critics. Double Q-learning reduces maximization bias by separating action selection from value evaluation \cite{hasselt2010double,van2016deep}, while TD3 extends this idea to continuous control through clipped double critics, delayed policy updates, and target-policy smoothing \cite{fujimoto2018td3}. PPO provides a useful on-policy contrast, where the critic is trained under a different update regime \cite{schulman2017ppo}. These methods improve critic reliability indirectly through targets, architectures, or optimization procedures.

A complementary route is the critic training process itself: whether critic complexity can be measured, related to performance, and controlled during training. The guiding hypothesis is Occam-style: among critics that capture the relevant value structure, a simpler critic may be more reliable, matching broader neural-network generalization arguments that use norm-based and spectral quantities as complexity proxies \cite{zhang2016understanding,bartlett2017spectrally}. This does not imply that lower complexity is always better. A critic that is too simple may underfit, while a critic that is too complex may overestimate, become unstable, or represent sharp value artifacts.

To make this hypothesis testable, critic complexity is measured using spectral effective-rank entropy. This metric summarizes how diffusely a critic layer uses its singular directions: high entropy means many directions contribute, while low entropy means the spectrum is concentrated in fewer dominant directions \cite{roy2007effective}. This quantity is then tracked during TD3 and PPO training, compared with return and Monte Carlo estimates of value-estimation bias, and directly controlled in separate experiments by adding a spectral-entropy penalty to the critic loss.

The main conclusion is deliberately narrow. Critic spectral complexity is measurable and controllable, and it is related to actor-critic performance in structured but task-dependent ways. Spectral-entropy regularization reliably reduces critic rank-like complexity and improves TD3/HalfCheetah-v4 performance in the tested setting, but the return benefit does not transfer cleanly across all tasks.

Concrete contributions:

\begin{itemize}
  \item Defines critic complexity as an explicit evaluation dimension for actor-critic reinforcement learning.
  \item Uses spectral effective-rank entropy as a computable rank-like measure of critic complexity.
  \item Provides observational evidence that critic complexity is systematically related to performance and bias, but not through a simple monotonic rule.
  \item Introduces a spectral-entropy regularizer for directly controlling critic complexity during critic training.
  \item Shows that the regularizer reliably reduces critic rank-like complexity and can improve TD3/HalfCheetah-v4 performance, while cross-task results remain mixed.
\end{itemize}
\section{Related Work}

Three strands of prior work are most relevant. First, value-estimation
bias is a standard concern in reinforcement learning. Double Q-learning and
Deep Double Q-learning reduce maximization bias by separating action
selection from value evaluation \cite{hasselt2010double,van2016deep}, and
TD3 adapts this motivation to continuous control with clipped double
critics, delayed policy updates, and target-policy smoothing
\cite{fujimoto2018td3}. PPO is included as an on-policy contrast
\cite{schulman2017ppo}, while SAC is relevant mainly as a policy-entropy
baseline: unlike SAC's maximum-entropy objective \cite{haarnoja2018sac},
the intervention regularizes the critic's weight spectrum rather than the
policy.

Second, neural-network generalization work has used norm-based and
spectral quantities as complexity proxies
\cite{zhang2016understanding,bartlett2017spectrally}.
Spectral normalization shows that singular-value control can also be a
practical training tool \cite{miyato2018spectral}. Effective rank is a
continuous alternative to exact matrix rank and has been studied as a
measure of effective dimensionality \cite{roy2007effective}.

Third, recent work connects complexity control to grokking and reasoning,
mostly in transformer or supervised compositional settings. Liu et al.
relate grokking to weight norm and loss-landscape mismatch
\cite{liu2023omnigrok}, while Zhang et al. show that initialization scale
and weight decay can steer transformers toward lower-complexity
reasoning-based solutions \cite{zhang2025complexity}. Musat gives a
complementary theoretical link between weight norm and Kolmogorov
complexity for fixed-precision looped or recursive transformer-style
models \cite{musat2026norm}. These papers motivate the view of complexity
as a controllable training variable; the experiments test that idea on
neural critics in actor-critic RL.

\section{Problem Formulation and Method}

The problem is to turn critic complexity from an informal intuition into a
quantity that can be computed, monitored, and intervened on during
actor-critic training. Before complexity can be measured in training, it
must be defined in a way that is computable from neural critic parameters
and activations. Let
$W \in \mathbb{R}^{m \times n}$ denote a weight matrix in the critic, and
let $\sigma_1 \geq \sigma_2 \geq \cdots \geq \sigma_r$ be its singular
values.

\subsection{Spectral Complexity Metric}

The spectral complexity metric is rank entropy. For
each critic weight matrix, define the normalized singular-value
distribution
\begin{equation}
  p_i = \frac{\sigma_i}{\sum_j \sigma_j}.
\end{equation}
The layer rank entropy is
\begin{equation}
  H(W) = -\sum_i p_i \log p_i,
\end{equation}
and the aggregate critic complexity metric is the average of this entropy
across critic layers. High entropy means that many singular directions
contribute meaningfully; low entropy means that the matrix is dominated by
a smaller number of directions. Effective-rank entropy is the primary complexity metric for three reasons.
\begin{enumerate}
  \item It is based on singular values, so it is easy to compute from critic
  weight matrices during training.
  \item The entropy scale is more interpretable than raw rank or norm
  values: researchers can read it as the spread of mass across spectral
  directions.
  \item In preliminary experiments, rank entropy was more correlated with task performance than other spectral metrics such as effective rank or stable rank.
\end{enumerate}

\subsection{Additional Metrics}

The analysis relates critic complexity to three outcome measurements. Bias
is the signed critic error under the current policy. At a checkpoint, the
policy is frozen, Monte Carlo rollouts estimate the return from sampled
states, and the critic prediction is compared to that return:
\begin{equation}
  \operatorname{Bias}(s) =
  Q_\theta(s,\pi(s)) - V_{\mathrm{MC}}(s).
\end{equation}
Positive values indicate overestimation, while negative values indicate
underestimation. \textbf{Return volatility} is the standard deviation of episodic
returns over the final 25 monitor episodes. \textbf{Bias volatility} is the standard
deviation of the checkpoint critic-bias estimates across sampled
evaluation states. These three quantities are compared against critic
effective-rank entropy at the run and checkpoint levels.

\subsection{Complexity-Control Intervention}

The control stage introduces one targeted intervention: spectral-entropy
regularization of the critic. The regularizer penalizes entropy in the
critic's singular-value distribution. For critic loss
$\mathcal{L}_{\mathrm{critic}}$, the modified objective is
\begin{equation}
  \mathcal{L}_{\mathrm{total}}
  =
  \mathcal{L}_{\mathrm{critic}}
  + \lambda_{\mathrm{ent}}
  \sum_{\ell \in \mathcal{C}} H(W_\ell),
\end{equation}
where $\mathcal{C}$ is the set of critic layers and
$\lambda_{\mathrm{ent}}$ is implemented as
\texttt{critic\_entropy\_coef}. Penalizing entropy encourages the critic to
concentrate its weight spectrum into fewer dominant directions, thereby
reducing rank entropy.

\section{Results}
\label{sec:results}

All experiments use PPO or TD3 actor-critic runs with critic complexity,
return, and checkpoint value-bias measurements logged during training.
TD3 uses separate actor and critic MLPs with two 256-unit hidden layers,
while PPO uses two 64-unit hidden layers for both the policy and value
networks.
Unless stated otherwise, analyses use completed runs only. Every run was
performed on a single H20 GPU.

\subsection{Observational Analysis}

\begin{figure}[t]
  \centering
  \includegraphics[width=\textwidth]{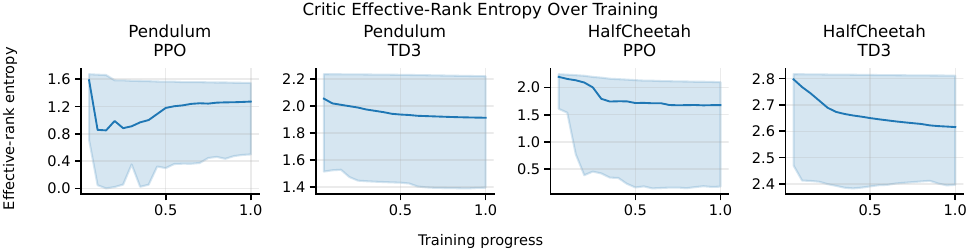}
  \caption{Evolution of critic effective-rank entropy over normalized
  training progress for non-controlled Pendulum-v1 and HalfCheetah-v4 runs.
  Each subplot summarizes one algorithm/task slice; the line shows the
  checkpoint median and the shaded band shows the interquartile range. The
  plot shows that critic effective-rank entropy generally decreases slightly during training, although with high variability. \textbf{Critic
  complexity is a dynamic training quantity, not only a final-run summary.}}
  \label{fig:entropy-over-training}
\end{figure}

\begin{figure}[t]
  \centering
  \includegraphics[width=\textwidth]{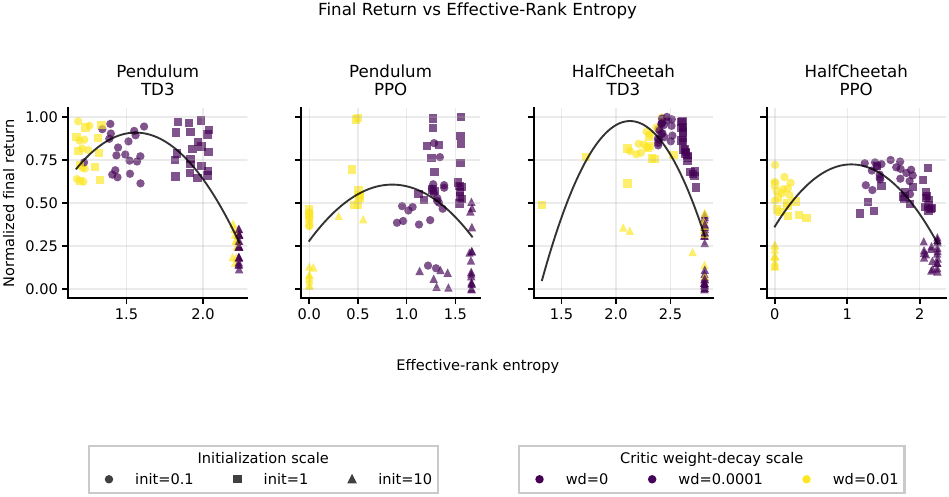}
  \caption{Run-level relationship between final return and critic
  effective-rank entropy for non-controlled Pendulum-v1 and HalfCheetah-v4
  runs. Ant-v4 and Walker2d-v4 are omitted to keep the diagnostic focused
  on the main baseline environments. \textbf{The best observed returns tend
  to occur at lower critic entropy, but the relationship depends on algorithm
  and task rather than following a universal monotonic rule.}}
  \label{fig:final-performance-complexity}
\end{figure}

The observational analysis asks whether critic complexity has a measurable
relationship to performance and bias under real training conditions.
It only uses non-controlled runs, meaning runs with no spectral-entropy
regularizer applied to the critic. The subset shown in
Figure~\ref{fig:final-performance-complexity} contains 360 PPO and TD3 runs
on Pendulum-v1 and HalfCheetah-v4. These runs vary seed
($\{0,1,2,3,4\}$), initialization scale ($\{0.1,1,10\}$), critic weight
decay ($\{0,10^{-4},10^{-2}\}$), and critic learning rate
($\{10^{-4},3\cdot 10^{-4}\}$).
Figure~\ref{fig:entropy-over-training} first checks whether critic
effective-rank entropy changes during training. Figure~\ref{fig:final-performance-complexity}
then shows the run-level relationship between final complexity and final
return. Together, the plots show that complexity and return are related in a
structured but not purely monotonic way: low rank entropy is not
automatically better, and high rank entropy is not automatically worse. Still,
the highest-return runs have lower effective-rank
entropy, suggesting that complexity is a meaningful signal for performance
even if it is not a simple monotonic one.

\begin{table}[t]
  \centering
  \caption{Run-level Spearman correlations for the non-controlled
  observational analysis. The x-axis variable is final critic
  effective-rank entropy; each row is one algorithm/task slice. Each row
  summarizes $n=90$ non-controlled runs from the observational sweep.
  \textbf{Entropy is most strongly associated with bias on Pendulum and with
  final return for TD3, showing that the signal is structured but task- and
  algorithm-dependent.}}
  \label{tab:heatmap-correlations}
  \begin{tabular}{lrrrrr}
    \toprule
    Algorithm / Task & Final return & Bias & Return volatility & Bias volatility \\
    \midrule
    PPO / Pendulum  & $0.03$ & $-0.75$ & $-0.21$ & $-0.45$ \\
    TD3 / Pendulum  & $-0.67$ & $-0.74$ & $0.52$ & $0.46$ \\
    PPO / HalfCheetah & $-0.21$ & $0.03$ & $-0.09$ & $-0.08$ \\
    TD3 / HalfCheetah & $-0.65$ & $-0.55$ & $-0.29$ & $0.16$ \\
    \bottomrule
  \end{tabular}
\end{table}

Table~\ref{tab:heatmap-correlations} gives a compact summary of the same
pattern. The strongest associations are task- and algorithm-dependent:
Pendulum shows a strong negative relationship between effective-rank
entropy and signed critic bias, while TD3 shows strong negative
relationships between entropy and final return on both baseline tasks.
These results should be read as descriptive rather than causal because the
relationships are heterogeneous across tasks and confounded by
hyperparameters. They motivate the intervention below: if complexity is
more than a passive correlate, directly changing it should alter training
dynamics.

\subsection{Complexity Control Intervention}

The main intervention subset contains 32 balanced runs: PPO and TD3 on
Pendulum-v1 and HalfCheetah-v4, using seeds 0 and 1 and entropy
coefficients $\{0,0.001,0.003,0.01\}$.
Table~\ref{tab:main-regularizer} reports the results. Each row averages the same two seeds, 0 and 1, after
de-duplicating repeated completed runs. Bold entries mark the best
coefficient within each algorithm/task group for that metric: higher final
return, lower volatility, smaller absolute final bias, and lower rank
entropy.

\begin{table}[t]
  \centering
  \caption{Main-task regularizer summary across PPO and TD3 on Pendulum-v1
  and HalfCheetah-v4. All rows use seeds 0 and 1; values are mean $\pm$
  SEM. Bold marks the best value within each algorithm/task block across
  entropy coefficients. \textbf{Entropy regularization reliably lowers TD3
  critic entropy, while the clearest performance gain appears for
  TD3/HalfCheetah-v4 at a moderate coefficient.}}
  \label{tab:main-regularizer}
  \resizebox{\textwidth}{!}{%
  \begin{tabular}{llrrrrr}
    \toprule
    Algorithm / Task & Coef. & Final-10 return & Return volatility & Final bias & Bias volatility & Rank entropy \\
    \midrule
    PPO / Pendulum & $0$ & \best{$-1537 \pm 90$} & $56.0 \pm 4.3$ & \best{$-104.2 \pm 15.2$} & \best{$44.3 \pm 14.3$} & $1.669 \pm 0.001$ \\
    PPO / Pendulum & $0.001$ & $-1542 \pm 82$ & $70.2 \pm 27.9$ & $-111.6 \pm 6.6$ & $49.4 \pm 19.9$ & $1.669 \pm 0.002$ \\
    PPO / Pendulum & $0.003$ & $-1557 \pm 65$ & \best{$45.3 \pm 3.1$} & $-112.3 \pm 10.1$ & $48.7 \pm 1.2$ & \best{$1.667 \pm 0.001$} \\
    PPO / Pendulum & $0.01$ & $-1540 \pm 84$ & $49.1 \pm 2.7$ & $-106.5 \pm 14.6$ & $55.5 \pm 0.1$ & $1.668 \pm 0.002$ \\
    \midrule
    TD3 / Pendulum & $0$ & \best{$-1334 \pm 51$} & \best{$250.3 \pm 7.2$} & $-226.7 \pm 17.8$ & $40.1 \pm 0.0$ & $2.234 \pm 0.000$ \\
    TD3 / Pendulum & $0.001$ & \best{$-1334 \pm 51$} & \best{$250.3 \pm 7.2$} & $-231.1 \pm 16.8$ & \best{$35.7 \pm 0.5$} & $2.057 \pm 0.009$ \\
    TD3 / Pendulum & $0.003$ & \best{$-1334 \pm 51$} & \best{$250.3 \pm 7.2$} & $-225.0 \pm 16.7$ & $41.0 \pm 2.5$ & $1.865 \pm 0.022$ \\
    TD3 / Pendulum & $0.01$ & \best{$-1334 \pm 51$} & \best{$250.3 \pm 7.2$} & \best{$-224.1 \pm 11.2$} & $40.8 \pm 0.4$ & \best{$1.761 \pm 0.020$} \\
    \midrule
    PPO / HalfCheetah & $0$ & $-623 \pm 47$ & $89.4 \pm 18.2$ & $-10.8 \pm 4.3$ & $29.8 \pm 1.8$ & $2.243 \pm 0.001$ \\
    PPO / HalfCheetah & $0.001$ & $-651 \pm 49$ & \best{$82.7 \pm 0.5$} & $-9.2 \pm 8.5$ & \best{$27.0 \pm 1.3$} & \best{$2.242 \pm 0.001$} \\
    PPO / HalfCheetah & $0.003$ & \best{$-620 \pm 46$} & $82.9 \pm 8.0$ & \best{$-4.8 \pm 7.5$} & $27.2 \pm 2.7$ & $2.244 \pm 0.001$ \\
    PPO / HalfCheetah & $0.01$ & $-641 \pm 26$ & $91.7 \pm 5.2$ & $-6.7 \pm 6.1$ & $29.7 \pm 2.7$ & $2.243 \pm 0.001$ \\
    \midrule
    TD3 / HalfCheetah & $0$ & $-789 \pm 235$ & $73.6 \pm 66.2$ & $-200.5 \pm 50.4$ & $17.1 \pm 6.9$ & $2.811 \pm 0.000$ \\
    TD3 / HalfCheetah & $0.001$ & \best{$-712 \pm 10$} & \best{$12.3 \pm 9.4$} & \best{$-126.4 \pm 27.4$} & \best{$6.3 \pm 0.9$} & $2.800 \pm 0.007$ \\
    TD3 / HalfCheetah & $0.003$ & $-798 \pm 113$ & $42.9 \pm 37.3$ & $-149.0 \pm 2.5$ & $11.8 \pm 5.7$ & $2.802 \pm 0.011$ \\
    TD3 / HalfCheetah & $0.01$ & $-738 \pm 255$ & $20.0 \pm 6.2$ & $-227.8 \pm 59.7$ & $13.4 \pm 6.7$ & \best{$2.742 \pm 0.018$} \\
    \bottomrule
  \end{tabular}
  }
\end{table}

Across the four main slices, the regularizer most clearly moves the
targeted metric for TD3: rank entropy decreases on Pendulum-v1 and
HalfCheetah-v4 as the coefficient increases. The performance effect is not
uniform. The strongest return improvement appears in
Table~\ref{tab:main-regularizer}: under the balanced seed-0/1 comparison,
TD3/HalfCheetah-v4 has the best final return at coefficient $0.001$, while
its lowest rank entropy occurs at coefficient $0.01$. The PPO rows show
much smaller rank-entropy changes and no comparable return effect.

The bias result is more subtle. Final signed bias does not improve
reliably across all four slices. The HalfCheetah TD3 volatility measures
are more directionally consistent: final return volatility and final bias
volatility both decrease under the best regularized setting. Therefore, the data do not support the simple story
that the regularizer improves return by reducing final bias. Instead, the
more defensible interpretation is that the regularizer changes critic
complexity and stability dynamics, and that this change is beneficial most
clearly for TD3/HalfCheetah-v4.

\noindent\textbf{Cross-task generalization.}
The cross-task experiment tests
$\texttt{critic\_entropy\_coef}=0.01$ on two additional TD3 tasks:
Walker2d-v4 and Ant-v4. To match Table~\ref{tab:main-regularizer}, each
condition uses seeds 0 and 1,
with the same non-entropy hyperparameters as the main intervention subset.

\begin{table}[t]
  \centering
  \caption{Cross-task regularizer summary. All rows use seeds 0 and 1;
  values are mean $\pm$ SEM. Bold marks the best value within each
  algorithm/task block across entropy coefficients, using the same columns
  and highlighting convention as Table~\ref{tab:main-regularizer}.
  \textbf{The regularizer still reduces critic entropy on Walker2d-v4 and
  Ant-v4, but this control effect does not translate into a consistent
  return improvement.}}
  \label{tab:cross-task}
  \resizebox{\textwidth}{!}{%
  \begin{tabular}{llrrrrr}
    \toprule
    Algorithm / Task & Coef. & Final-10 return & Return volatility & Final bias & Bias volatility & Rank entropy \\
    \midrule
    TD3 / Walker2d & $0$ & \best{$-7.1 \pm 0.51$} & \best{$0.092 \pm 0.002$} & $-13.3 \pm 10.6$ & \best{$0.729 \pm 0.083$} & $2.816 \pm 0.000$ \\
    TD3 / Walker2d & $0.01$ & $-10.8 \pm 7.06$ & $0.243 \pm 0.153$ & \best{$-11.7 \pm 11.0$} & $1.166 \pm 0.053$ & \best{$2.573 \pm 0.086$} \\
    \midrule
    TD3 / Ant & $0$ & $-2702.5 \pm 0.14$ & \best{$4.46 \pm 0.13$} & \best{$-175.6 \pm 52.1$} & $21.8 \pm 3.2$ & $2.953 \pm 0.001$ \\
    TD3 / Ant & $0.01$ & \best{$-2702.0 \pm 0.66$} & $4.69 \pm 0.24$ & $-249.1 \pm 142.1$ & \best{$16.9 \pm 5.3$} & \best{$2.901 \pm 0.019$} \\
    \bottomrule
  \end{tabular}
  }
\end{table}

Table~\ref{tab:cross-task} shows that on Walker2d-v4, rank entropy drops
from about $2.82$ to $2.57$ at
coefficient $0.01$, but return and both volatility measures do not improve
under the seed-0/1 comparison. Signed final bias becomes slightly less
negative, but the uncertainty is large. On Ant-v4, the rank-entropy
reduction is smaller but still visible; final return is essentially
unchanged, return volatility is slightly worse, and final bias volatility
decreases. The cross-task result therefore separates two claims.
Spectral-entropy regularization is a robust control knob for critic
complexity. It is not, at the tested coefficient, a task-general
return-improvement method.

\section{Conclusion}

Critic complexity in actor-critic reinforcement learning is studied through
a three-part arc: gauging, measuring, and controlling. The gauging stage
defined computable complexity metrics based on singular-value spectra and
local input sensitivity. The measuring stage built a pipeline that tracks
these metrics during full RL training and relates them to return and Monte
Carlo bias estimates. The controlling stage introduced a targeted
spectral-entropy regularizer for the critic.

The central finding is that critic spectral complexity is actionable.
Rank entropy can be measured over time, it shows structured training
dynamics, and it can be directly reduced by a regularizer. On
TD3/HalfCheetah-v4, this reduction coincides with a large return
improvement. On
Walker2d-v4 and Ant-v4, complexity reduction still occurs, but the return
benefit does not transfer cleanly.

The main limitation is scope. The experiments do not establish that the
effect is algorithm-general, because a broader cross-algorithm campaign was
not run. They also do not establish full hyperparameter robustness, because the
robustness sweep over initialization scale and weight decay was not run.
The final claim is therefore precise: spectral entropy regularization is an
effective way to control critic rank-entropy dynamics, and those dynamics reveal
training behavior that is not visible from return curves alone.

\end{document}